\documentclass[twocolumn,10pt]{article}
\usepackage[twocolumn,textwidth=19cm,columnsep=.5cm, lines=57]{geometry}
\usepackage{listings}
\usepackage{setspace}
\usepackage{amsthm,amsmath,amssymb}
\usepackage{thm-restate}
\usepackage{etoolbox}
\usepackage{nicefrac}
\usepackage{tikz}
\usepackage{complexity}
\usepackage{todonotes}

\usepackage{natbib}
\setcitestyle{authoryear,open={(},close={)}}

\usetikzlibrary{shapes.geometric}
\usetikzlibrary{positioning}
\tikzstyle{arg}=[draw,circle,fill=gray!15,inner sep=1pt,minimum size=.5cm]

\usepackage{todonotes}
\usepackage{times}
\usepackage{soul}
\usepackage{url}
\usepackage[hidelinks]{hyperref}
\usepackage[utf8]{inputenc}
\usepackage[small]{caption}
\usepackage{graphicx}
\usepackage{amsmath,amsfonts}
\usepackage{multirow}
\usepackage{amsthm}
\usepackage{booktabs}
\usepackage{algorithm}
\usepackage{algorithmic}
\usepackage{xcolor}
\newcommand{\PS}[1]{\textcolor{black}{#1}}

\newtheorem{example}{Example}

\newtheorem{definition}{Definition}

\title{PeerArg: Argumentative Peer Review with LLMs}

\author{
\textbf{Purin Sukpanichnant,
Anna Rapberger,
Francesca Toni}\\[0.5ex]
Imperial College London, United Kingdom
\\[1ex]
\{ps1620, a.rapberger, ft\}@imperial.ac.uk
}
\date{}

\begin{document}

\maketitle

\begin{abstract}
  Peer review is an essential process to determine the quality of  papers submitted to scientific conferences or journals. However, it is subjective and prone to biases. Several studies have been conducted to apply techniques from NLP to 
  support peer review, but they are based on black-box techniques and their outputs are difficult to interpret and trust. In this paper, we propose a novel 
  pipeline to 
  support and understand the reviewing and decision-making processes of peer review: the PeerArg system combining LLMs with methods from knowledge representation. PeerArg takes in input a set of reviews for a paper and outputs the paper acceptance prediction. We evaluate the performance of 
  the PeerArg pipeline on three different datasets, in comparison with a novel end-2-end LLM that uses few-shot learning to predict paper acceptance given reviews. The results indicate that 
  the end-2-end LLM is capable of predicting paper acceptance from reviews, but 
  a variant of the PeerArg pipeline outperforms this LLM.
\end{abstract}

\section{Introduction}

Peer review is a process where work is examined and evaluated by a group of people with expertise in the relevant field. The process is crucial to ensure the quality of the work. It has been adopted by many conferences and journals, to ensure the published papers are of adequate quality. Peer review is hence a core component of the progress in several academic areas.
Nevertheless, fairness is the main weakness of the process. Peer review involves lots of discussion and evaluation from human reviewers, which are subjective and are prone to biases and irrationality. For example, the reviewers may be more likely to accept a paper whose results agree with what they believe, known as confirmation bias \citep{CONFIRMATION-BIAS}. Another bias is first-impression bias, where the initial impressions of the document (e.g. typographical layout \citep{FIRST-IMPRESSION-BIAS}) affect the entire judgement.

There has been an emerging trend to study how to apply techniques from Natural Language Processing (NLP) to improve 
the peer review process. \citeauthor{LLM-AS-REVIEWER}~(\citeyear{LLM-AS-REVIEWER}) study review generation and found that the generated reviews consider more aspects of a paper than human reviewers. Another example is given by \citeauthor{STATCHECK}~(\citeyear{STATCHECK}), with a tool that checks for statistical inconsistencies in papers.
Review understanding is another application of NLP in peer review. This could lessen the burden for meta-reviewers or conference chairs who read the reviews before deciding whether to accept a paper or not. For example, \citeauthor{APCS}~(\citeyear{APCS}) propose methods to generate pros and cons summary given reviews. \citeauthor{METAGEN}~(\citeyear{METAGEN}) present a system to generate a meta-review from reviews.
Most of the studies for review understanding used deep learning models as their backbone. Even though they have impressive performances, the black-box nature of the models makes it difficult to 
understand the rationale behind the results and trust the models.
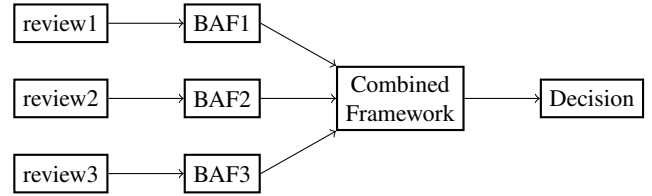
\begin{figure}[t]
    \centering
    \begin{tikzpicture}[
    squarednode/.style={rectangle, draw=black, thick, minimum size=5mm, font=\small},
    ]
    \node[squarednode][align=center] (r1) at (0,0) {review1};
    \node[squarednode][align=center] (r2) at (0,-1)  {review2};
    \node[squarednode][align=center] (r3) at (0,-2) {review3};

    \node[squarednode][align=center] (af1) [right=of r1] {BAF1};
    \node[squarednode][align=center] (af2) [right=of r2] {BAF2};
    \node[squarednode][align=center] (af3) [right=of r3] {BAF3};

    \node[squarednode][align=center] (cf) [right=of af2] {Combined \\ Framework};

    \node[squarednode][align=center] (dec) [right=of cf] {Decision};
    
    \draw[->] (r1.east) -- (af1.west);
    \draw[->] (r2.east) -- (af2.west);
    \draw[->] (r3.east) -- (af3.west);
    \draw[->] (af1.east) -- (cf.north west);
    \draw[->] (af2.east) -- (cf.west);
    \draw[->] (af3.east) -- (cf.south west);
    \draw[->] (cf.east) -- (dec.west);
    \end{tikzpicture}
    
    \caption{Overview of the PeerArg pipeline. Firstly, a bipolar argumentation framework BAF $i$ is extracted from review $i$. Then, the frameworks are combined. 
    The final decision is drawn from the combined framework.} 
    \label{fig:peerarg-contribution-diagram}
\end{figure}

In this paper, we propose a new technique to address these shortcomings. We use symbolic AI methods, specifically methods from computational argumentation~\citep{DUNG-AAF,BAF,QBAF}, to enhance review understanding and to assess the alignment between paper acceptance decision-making process and the reviews. 
Computational argumentation studies how to model the way humans build and use arguments so that it can be represented in a machine \citep{TOWARDS-ARTIFICIAL-ARGUMENTATION}. 
Whereas existing argumentation studies in peer review focus on the structure of arguments~\citep{HUA-2019-ARGUMENT,AM-ANALYSIS-PEER-REVIEW}, our work uses a form of bipolar argumentation frameworks~\citep{BAF}, adding a support relation to abstract argumentation~\citep{DUNG-AAF}, to model both the reviews and the review aggregation process.
Our main contribution is \emph{PeerArg}, a novel framework for transparent review aggregation. 
PeerArg  predicts the acceptance status of the paper, as illustrated in Figure \ref{fig:peerarg-contribution-diagram}. 

As part of this paper, we furthermore propose a few-shot learning end-2-end LLM taking in reviews of a paper and predict the paper acceptance, inspired from \citeauthor{LLM-AS-REVIEWER}~(\citeyear{LLM-AS-REVIEWER}) and \citeauthor{LLM-IN-RBAM}~(\citeyear{LLM-IN-RBAM}).
We evaluate PeerArg and the end-2-end LLM on three review datasets: two conference review datasets, and a journal review dataset. 
Our empirical studies indicate that enhancing LLMs with methods from computational argumentation has beneficial effects.
The results show that the LLM can be used to predict the paper acceptance from reviews, and with a certain hyperparameters combination, the PeerArg pipeline outperforms the LLM in all the datasets. 

\section{Related Work}

Methods for improving the interpretation and understanding of reviews using AI have received significant attention in recent years. 
\paragraph{Understanding Reviews with AI}
Several components of peer review have the potential to be improved using AI, such as pre-review screening, matching papers with reviewers, or review generation. 
One direction to enhance the reviewing process is review summarisation. 
\citeauthor{APCS}~(\citeyear{APCS}) propose methods to generate a pros and cons summary of given reviews of a paper.
\citeauthor{METAGEN}~(\citeyear{APCS}) propose a tool called \emph{MetaGen} to generate a meta-review from given reviews.

Conferences 
give specific guidelines concerning with aspects they expect the reviewers to consider
. 
Several studies have been done around accurately predicting and understanding these scores as this enhances the trustworthiness of the reviews.
An example is PeerRead \citep{PEERREAD}, the first public dataset of scientific peer reviews with corresponding decisions and aspect scores, i.e., the ratings the reviewers gave for each of the aspects. This dataset was used in \citeauthor{MULTI-TASK-PEER-REVIEW-SCORE-PREDICTION}~(\citeyear{MULTI-TASK-PEER-REVIEW-SCORE-PREDICTION}) to evaluate the proposed multi-task framework for peer-review aspect score predictions. \citeauthor{ASPECT-BASED-SENTIMENT-ANALYSIS}~(\citeyear{ASPECT-BASED-SENTIMENT-ANALYSIS}) performed an aspect-based sentiment analysis and determined that there has been a correlation between the distribution of aspect-based sentiments and the acceptance decision of papers.
Recent studies are shifting to the use of a Large Language Model (LLM). \citeauthor{LLM-AS-REVIEWER}~(\citeyear{LLM-AS-REVIEWER}) evaluated two LLMs, GPT-3.5 and GPT-4, on the review scores prediction and review generation tasks. 
The results indicated that LLMs can infer aspect scores given a review; however, their performances were inadequate when given a paper.

\paragraph{Argumentation in Peer Review} \label{sec:arg-peer-review}
Methods from computational argumentation have rarely been applied in peer review applications. 
Notable exceptions are the works by~\citeauthor{HUA-2019-ARGUMENT}~(\citeyear{HUA-2019-ARGUMENT}) and \citeauthor{AM-ANALYSIS-PEER-REVIEW}~(\citeyear{AM-ANALYSIS-PEER-REVIEW}). However, in contrast to our work, neither consider the relations between arguments.

\citeauthor{HUA-2019-ARGUMENT}~(\citeyear{HUA-2019-ARGUMENT}) applied argumentation to understand the content and structure of peer reviews. They detect argumentative text in a review and classify it into one of the following types: evaluation, request, fact, reference, or quote. The authors then analysed and compared argumentation in reviews across several ML and NLP conferences. The results show that there have been some discrepancies in the argumentation trend across different conferences. For example, ACL and NeurIPS tend to contain most arguments, strong reject/accept reviews tend to have fewer arguments.

Another proposal to 
was made in \citeauthor{AM-ANALYSIS-PEER-REVIEW}~(\citeyear{AM-ANALYSIS-PEER-REVIEW}). The authors aimed to extract the most relevant arguments from a review and evaluated its effect towards the paper acceptance decision. The experiment empirically indicated that correct decisions can be made by using merely half of the review.

\section{Preliminaries}
We recall bipolar argumentation~\citep{BAF} and few-shot learning for LLMs~\citep{FEW-SHOT-LEARNING}.

\subsection{Bipolar Argumentation}
In bipolar argumentation frameworks~\citep{BAF}, arguments are abstract entities; relations between them are either supports or attacks.  We define them as follows.

\begin{definition}
    \label{def:baf}
    A \emph{bipolar argumentation framework} (BAF) is a tuple $\langle X, Att, Supp \rangle$ where $X$ is a finite set of arguments and $Att, Supp \subseteq X \times X$ are attack and support relations between arguments. An argument $a \in X$ attacks an argument $b \in X$ if and only if $(a, b) \in Att$. Similarly, an argument $a \in X$ supports an argument $b \in X$ if and only if $(a, b) \in Supp$.
\end{definition}

We furthermore consider quantitative bipolar argumentation frameworks (QBAF) \citep{QBAF}.

\begin{definition}
\label{def:qbaf}
    A \emph{quantitative BAF (QBAF)} is a tuple $\langle X, Att, Supp, \beta \rangle$ over range $D=[0, 1]$
    where $\langle X,Att,Supp\rangle$ is a BAF and $\beta: X \rightarrow D$ is a total function that assigns a base score to each argument. 
\end{definition}

By $A(a) = \{b|(b, a) \in Att\}$ we denote the attackers, by $S(a) = \{b|(b, a) \in Supp\}$ the supporters of an argument $a$.

A semantics $\sigma_Q: X \rightarrow D$ for a QBAF $Q$ determines the final strength of each argument. In this work, we use the DF-QuAD semantics \citep{DF-QUAD} and MLP-based semantics \citep{NN-to-QBAF}.

\begin{definition}
\label{def:df-quad-semantics}
    Let $Q=\langle X, Att, Supp, \beta \rangle$ be a QBAF over $D=[0, 1]$. 
    Let $\delta: D^{*} \rightarrow D$ denote the strength aggregation function,\footnote{Here, $D^*$ is the set of all sequences of elements of $D$.} such that, for $T = (v_1,\ldots,v_n) \in D^{*}$:
    \begin{align*}
        & \text{if n = 0: } \delta(T) = 0; \\
        & \text{if n = 1: } \delta(T) = v_1; \\
        & \text{if n = 2: } \delta(T) = f(v_1, v_2); \\
        & \text{if n $>$ 2: } \delta(T) = f(\delta(v_1,\ldots,v_{n-1}), v_n)
    \end{align*}
    where,  $f(x, y) = x + (1 - x) \cdot y = x + y - x \cdot y$, $x, y \in D$.

    Let $\varphi: D \times D \times D \rightarrow D$ denote the influence function, where for $v_0, v_a, v_s \in D$:
    \[
        \varphi(v_0, v_a, v_s) = 
        \begin{cases} 
            v_0 - v_0 \cdot |v_s - v_a| & \text{if } v_a \geq v_s \\
            v_0 + (1 - v_0) \cdot |v_s - v_a| & \text{if } v_a < v_s 
        \end{cases}
    \].

    For any $a\in X$, the DF-QuAD semantics is defined by
    \[
         \sigma_{DF-QuAD}(a) = \varphi(\beta(a), \delta(\sigma(A(a))), \delta(\sigma(S(a))))
    \]
    s.t.\ $\sigma(A(a)) = (\sigma(a_1),\ldots,\sigma(a_n))$ where $(a_1,\ldots,a_n)$ is an arbitrary permutation of the $(n\geq0)$ attackers in A(a), and $\sigma(S(a)) = (\sigma(s_1),\ldots,\sigma(s_m))$ where $(s_1,\ldots,s_m)$ is an arbitrary permutation of the $(m\geq0)$ supporters in S(a).
\end{definition}
\PS{For each argument, DF-QuAD semantics aggregates the strengths of its attackers and supporters, and determines how both aggregates influence the base score of the argument.}

\begin{example}
\label{example:qbaf-df-quad}
An example of a QBAF is as illustrated below, with a set of arguments $X = \{a, b, c, d\}$, attack relation $Att = \{(b, c), (c, d)\}$, support relation $Supp = \{(a, c)\}$, base score function $\beta(a) = 0.5$, $\beta(b) = 0.4$, $\beta(c) = 0.2$, and $\beta(d) = 0.7$.
The evaluation under DF-QuAD semantics yields the final strength function $\sigma(a) = 0.5$, $\sigma(b) = 0.4$, $\sigma(c) = 0.28$, and $\sigma(d) = 0.504$.
The base score and the final strength of the arguments are written in brackets.

\begin{center}
    \begin{tikzpicture}[
    squarednode/.style={rectangle, draw=black, thick, minimum size=5mm, font=\small},
    ]
    \node[squarednode][align=center] (a) {\textbf{a}: "My dog is cold."\\(0.5: 0.5)};
    \node[squarednode][align=center] (b) [below=of a] {\textbf{b}: "The weather \\ is hot." \\(0.4: 0.4)};
    \node[squarednode][align=center] (c) [right=of a] {\textbf{c}: "The heater \\ is on."\\(0.2: 0.28)};
    \node[squarednode][align=center] (d) [below=of c] {\textbf{d}: "Energy bill on\\ the heater is low."\\(0.7: 0.504)};
    
    \draw[->] (a.east) -- (c.west) node[midway, above] {\textcolor{green}{+}};
    \draw[->] (b.east) -- (c.south west) node[midway, above] {\textcolor{red}{\Large -}};
    \draw[->] (c.south) -- (d.north) node[midway, right] {\textcolor{red}{\Large -}};
    \end{tikzpicture}
\end{center}
\end{example}

\begin{definition}
\label{def:mlp-based-semantics}
    Let $Q=\langle X, Att, Supp, \beta \rangle$ be a QBAF over $D=[0, 1]$.
    The MLP-based semantics is defined by
    \[  
        \sigma_{MLP}(a) = 
        \begin{cases} 
            \lim_{k\to\infty} s_{a}^{(k)} & \text{if the limit exists} \\
            \perp & \text{otherwise} 
        \end{cases}
    \]
    for any argument a $\in X$, where $s_{a}^{(k)}$ is defined using a two-step process \citep{MODULAR-SEMANTICS}. Starting with $s_{a}^{(0)} = \beta(a)$ (i.e. a base score of that argument), we then iterate over two steps:
    \[
        \emph{Aggregation: } 
        \alpha_{a}^{(i+1)} = \Sigma_{(b, a) \in Supp} s_{b}^{(i)} - \Sigma_{(b, a) \in Att} s_{b}^{(i)}
    \]
    \[
        \emph{Influence: } 
        s_{a}^{(i+1)} = \varphi(\varphi^{-1}(\beta(a)) + \alpha_{a}^{(i+1)})
    \]
    where $\varphi: \mathbb{R} \rightarrow D$ is strictly monotonically increasing.
\end{definition}

\PS{MLP-based semantics treats a QBAF as a multi-layer perceptron (MLP), where edges from attacking/supporting arguments have weights -1 and +1 respectively. Strengths are calculated similar to a forward pass in an MLP.}

\begin{example}
\label{example:qbaf-mlp-based}
Let us head back to the QBAF from Example~\ref{example:qbaf-df-quad}.
Assuming $\varphi$ is the sigmoid function, the evaluation under MLP-based semantics yields the final strength function $\sigma(a) = 0.5$, $\sigma(b) = 0.4$, $\sigma(c) = 0.216$, and $\sigma(d) = 0.653$.
\end{example}

\subsection{Few-Shot Learning for LLMs}

Large language models (LLMs) are the pretrained transformer models \citep{TRANSFORMER} that take in a text input and generate the output text. LLMs are useful for multiple tasks such as machine translation, text summarisation, and text generation.
Due to the extremely large number of parameters in the LLMs e.g. 175 billion parameters for GPT-3 \citep{FEW-SHOT-LEARNING}, fine-tuning them for a particular task requires a large amount of computing resources and time.

Few-shot learning \citep{FEW-SHOT-LEARNING} is a way to resolve such constraints. The method works by providing the LLMs some examples of the task we expect them to perform, and let the LLMs learn to produce the result in a similar format as the provided examples for our inputs. This learning is done without fine-tuning, utilising the general knowledge the LLMs have acquired during their pre-training period. Examples are called \emph{primer} and the input is called \emph{prompt}, as in \citep{LLM-IN-RBAM}. 

\section{Paper acceptance by end-to-end LLMs} \label{sec:end-2-end-llm}

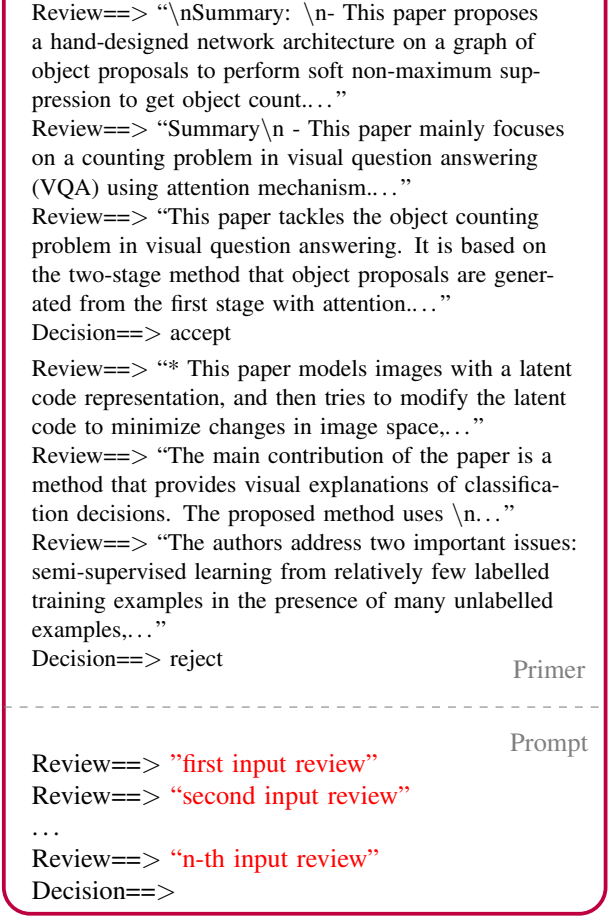
\begin{figure}[t]
    \centering
    \begin{tikzpicture}
        \draw[rounded corners=10pt, line width=0.5mm, purple] (0, 0) rectangle (8, 12.5);

        \draw[dashed, gray] (0, 2.75) -- (8, 2.75);

        \node[align=left] at (7.25, 3.25) {\textcolor{gray}{Primer}};
        \node[align=left] at (7.25, 2.25) {\textcolor{gray}{Prompt}};

        \node[align=left, text width=7.25cm, font=\small, below right] at (0.25, 12.25) {
            Review==\textgreater\ “\textbackslash nSummary: \textbackslash n- This paper proposes a hand-designed network architecture on a graph of object proposals to perform soft non-maximum suppression to get object count.…” \\
            Review==\textgreater\ “Summary\textbackslash n - This paper mainly focuses on a counting problem in visual question answering (VQA) using attention mechanism.…” \\
            Review==\textgreater\ “This paper tackles the object counting problem in visual question answering. It is based on the two-stage method that object proposals are generated from the first stage with attention.…” \\
            Decision==\textgreater\ accept
        };

        \node[align=left, text width=7.25cm, font=\small, below right] at (0.25, 7.5) {
            Review==\textgreater\ “* This paper models images with a latent code representation, and then tries to modify the latent code to minimize changes in image space,…” \\
            Review==\textgreater\ “The main contribution of the paper is a method that provides visual explanations of classification decisions. The proposed method uses \textbackslash n…” \\
            Review==\textgreater\ “The authors address two important issues: semi-supervised learning from relatively few labelled training examples in the presence of many unlabelled examples,…” \\
            Decision==\textgreater\ reject
        };

        \node[align=left, text width=7.25cm, below right] at (0.25, 2.25) {
            Review==\textgreater\ \textcolor{red}{"first input review”} \\
            Review==\textgreater\ \textcolor{red}{“second input review”} \\
            … \\
            Review==\textgreater\ \textcolor{red}{“n-th input review”} \\
            Decision==\textgreater\ 
        };
        
    \end{tikzpicture}
    \caption{End-2-End LLM Input Template (the sample reviews are truncated due to lack of space)}
    \label{fig:llm-e2e-template}
\end{figure}
\begin{figure*}[t]
    \centering
    \includegraphics[width=1.0\linewidth, keepaspectratio]{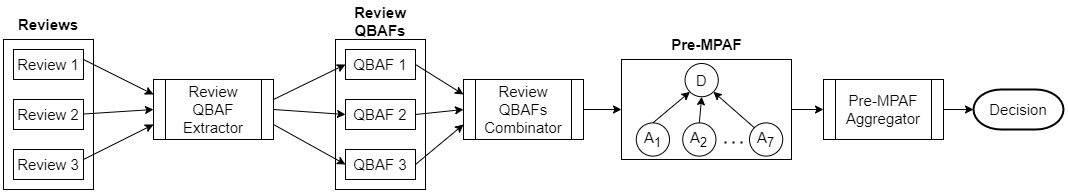}
    \caption{PeerArg pipeline diagram}
    \label{fig:peerarg-pipeline}
\end{figure*}
In this section, we propose an LLM that applies the few-shot learning methodology to classify a paper acceptance given the reviews of the paper, called \emph{end-2-end LLM}.\footnote{\url{https://gitlab.doc.ic.ac.uk/ps1620/peerarg/-/tree/master/llm_e2e?ref_type=heads}}

The end-2-end LLM uses a quantised 4-bit pretrained Mistral-7B-0.1 from Mistral AI\footnote{https://huggingface.co/mistralai/Mistral-7B-v0.1} as a pretrained LLM. The template of an input prompt given to the LLM is inspired from \citep{LLM-IN-RBAM}, consisting of a primer and a prompt. The primer consists of reviews of four papers each with the final decision, two accepts and two rejects, taken from~\citep{FIRST_E2E_LLM_SAMPLE,SECOND_E2E_LLM_SAMPLE,THIRD_E2E_LLM_SAMPLE,FOURTH_E2E_LLM_SAMPLE}. The prompt is a set of reviews of the paper we want to predict, with no labels. 

The partial template for the primer \& prompt input to LLM is shown in Figure \ref{fig:llm-e2e-template} consisting of two reviews-decision samples. Each review starts with \emph{Review==$>$} placeholder followed by the raw review in double quotation mark and a newline to separate each review. Each set of reviews is then followed by the decision which starts with \emph{Decision==$> $} and the accept/reject decision. These samples are the primer of our input to the LLM.

The prompt of the LLM input is in a review-decision format similar to the samples in the primer. The only difference is that there is no decision attached to the \emph{Decision==$>$} placeholder. This is to let the LLM predict the outcome.

\section{Peer-Review Enhanced with KR}
\label{sec:peerarg-details}

One downside of the end-2-end LLM is its black-box nature. The model only returns the final result without intermediary steps which makes it difficult to rationalise with. To resolve this issue, we incorporate knowledge representation into the process. In this section, we present the PeerArg pipeline\footnote{Repository: https://gitlab.doc.ic.ac.uk/ps1620/peerarg}. Given the reviews of a paper, the pipeline represents each of them as an argumentation framework, combining these frameworks into a single framework, and aggregating to determine the paper acceptance.

The pipeline diagram is as illustrated in Figure \ref{fig:peerarg-pipeline}, assuming that there are three reviewers for a paper. The pipeline consists of three main steps: \emph{Review QBAF extraction}, \emph{Review QBAFs combination}, and \emph{Pre-MPAF aggregation}.

\subsection{Review QBAF Extractor} \label{subsec:peerarg-review-qbaf-extractor}

The first step is to extract an argumentation framework from each review. 
Reviewers are typically encouraged to assess the quality of a paper under various aspects, each of which may affect the decision more/less than others, and provide justification for their impression of the paper w.r.t. these aspects.
Following this idea, we can treat aspects as arguments (called \emph{aspect arguments}) that attack or support the decision (the so-called \emph{decision argument}). Each sentence in a review is associated with different aspects, and so can be considered as an argument (called \emph{text argument}). As a result, each review can be considered as a three-level QBAF consisting of text and aspect arguments, and decision argument.

There are several different aspects that conferences expect reviewers to consider. These vary for different conferences. In this paper, we focus on the aspects provided by the ACL 2016 conference, namely \emph{appropriateness}, \emph{clarity}, \emph{novelty}, \emph{empirical and theoretical soundness}, \emph{meaningful comparison}, \emph{substance}, and \emph{impact}.
For our framework, we introduce an argument for each of these aspects; 
abbreviated by APR, CLA, NOV, EMP, CMP, SUB, and IMP, respectively.
Below, we introduce our review QBAFs.

\begin{definition}
\label{def:review-qbaf}
    A review QBAF is a tuple $\langle X, Att, Supp, \beta \rangle$ over $D = [0, 1]$, where $X = T \cup A \cup \{Decision\}$ with T as the set of text arguments, A = \{APR, CLA, NOV, EMP, CMP, SUB, IMP\} as the set of aspect arguments, and Decision as the decision argument. The relations $Att, Supp \subseteq (T \times A) \cup (A \times {Decision})$ are such that Att and Supp are disjoint. A semantics $\sigma: X \rightarrow D$ assigns a strength to each argument.
\end{definition}

\begin{example}\label{ex:review-qbaf-example}
An example of a review QBAF is 
given below. 
\begin{center}
      \includegraphics[width=1.0\linewidth, keepaspectratio]{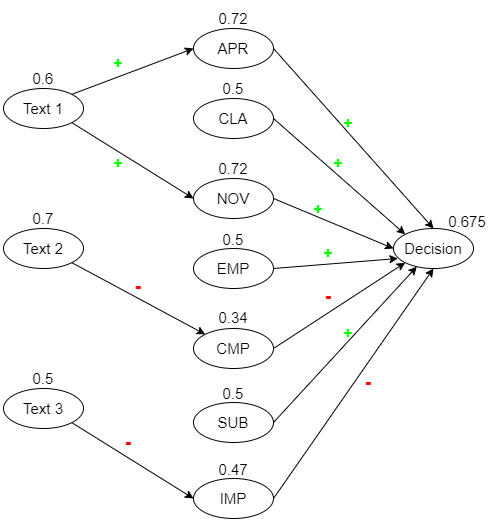}  
\end{center}
This review QBAF has three text arguments, called \emph{Text 1}, \emph{Text 2} and \emph{Text 3} which attack/support different aspect arguments; e.g., \emph{Text 1} supports APR (appropriateness) as well as NOV (novelty of the work). Each argument is annotated with the final strength, depicted above it. Edges with minus are attacks and edges with plus are supports. 
\end{example}

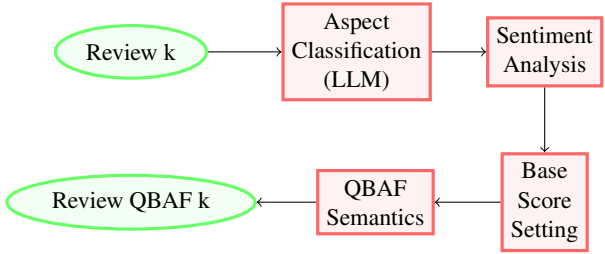
\begin{figure}[t]
    \centering
    
    \begin{tikzpicture}[
    ellipsenode/.style={ellipse, draw=green!60, fill=green!5, very thick, minimum size=7mm, font=\small},
    squarednode/.style={rectangle, draw=red!60, fill=red!5, very thick, minimum size=5mm, font=\small},
    ]
    
    \node[ellipsenode][align=center] (review) at (-0.5,0) {Review k};
    \node[squarednode][align=center] (aspectClassification) at (2.5,0) {Aspect\\Classification\\(LLM)};
    \node[squarednode][align=center] (sentimentAnalysis) at (5,0) {Sentiment\\Analysis};
    \node[squarednode][align=center] (baseScoreSetting) at (5,-2) {Base\\Score\\Setting};
    \node[squarednode][align=center] (qbafSemantics) at (2.75,-2) {QBAF\\Semantics};
    \node[ellipsenode][align=center] (reviewQbaf) at (-0.5,-2) {Review QBAF k};
    
    \draw[->] (review.east) -- (aspectClassification.west);
    \draw[->] (aspectClassification.east) -- (sentimentAnalysis.west);
    \draw[->] (sentimentAnalysis.south) -- (baseScoreSetting.north);
    \draw[->] (baseScoreSetting.west) -- (qbafSemantics.east);
    \draw[->] (qbafSemantics.west) -- (reviewQbaf.east);
    \end{tikzpicture}
    
    \caption{The process to obtain a review QBAF from a review.}
    \label{fig:review-2-review-qbaf}
\end{figure}

To obtain a review QBAF from a review, there are four main steps, as illustrated in Figure \ref{fig:review-2-review-qbaf}.
Initially, we start with \textbf{aspect classification} where each sentence in the review is classified which aspects it belongs to. In this paper, we use a few-shot learning LLM for aspect classification. 
The next step is \textbf{sentiment analysis} to determine if the sentence is positive/negative towards the aspects. 
A graph structure of the review QBAF is obtained at this step.
Then, the third step is the \textbf{base score setting} where the base scores are set for all the arguments. Finally, \textbf{QBAF semantics} is applied to calculate the final strength of each argument.

\begin{figure}[t]
    \includegraphics[width=1.0\linewidth, keepaspectratio]{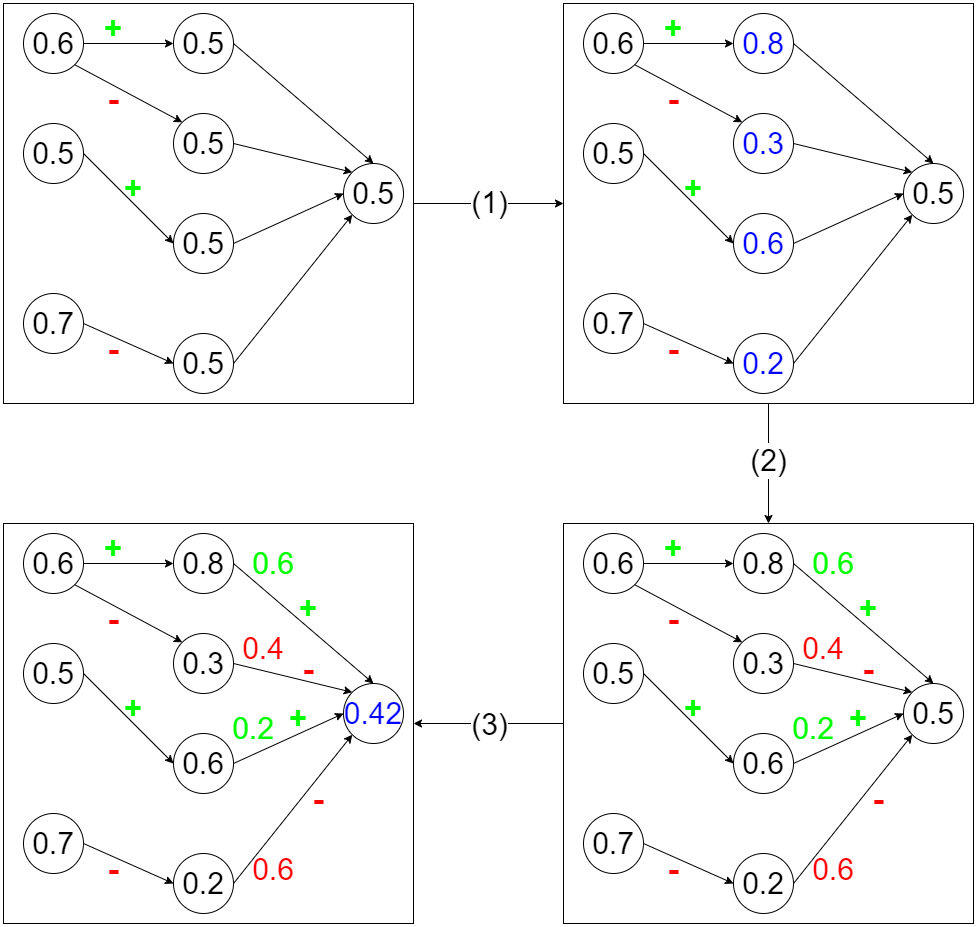}
    \caption{From an incomplete QBAF to a complete QBAF, with edges with minus and plus being attacks and supports respectively. (1) First, we apply QBAF semantics to get strengths for aspect arguments. (2) From the aspect arguments' strengths, we determine their relation to the decision argument; and calculate their scores (red and green numbers). (3) We apply QBAF semantics to get the final strength of the decision argument.}
    \label{fig:get-review-qbaf}
\end{figure}

Initially, all edges between aspect arguments and the decision argument are undecided, because each aspect may support or be against the paper being accepted, depending on it's strength.
Figure \ref{fig:get-review-qbaf} illustrates the process to determine the relations of the undecided edges of the incomplete QBAF of a review, up to the calculation of the strength of the decision argument.

To determine the relations from the aspect arguments to the decision argument, we first apply a (gradual) semantics to calculate the strengths of the aspect arguments. 
These strengths indicate how supportive the aspects are towards paper acceptance. A strength of 0 means a paper is very poor on this aspect, in contrast, a strength of 1 means the paper is excellent on this aspect. We set 0.5 as a midpoint. An aspect argument with strength below 0.5 is attacking; otherwise, it is supporting the decision argument.

\PS{The further the strength is away from 0.5, the stronger attacker/supporter an aspect argument is. To incorporate this idea, we reflect the strength around 0.5 before scaling up by a factor of 2 so that the strength is still in the range [0, 1].}
Formally, for an aspect argument $a$ of strength $s$, we define 
$$\beta(a)= 2\cdot|s - 0.5|.$$ Once these strengths (towards the decision argument) of aspect arguments are calculated, the semantics is then applied to calculate the final strength of the decision argument.

We depict the original strengths of the aspect arguments 
inside nodes in the bottom-left box in Figure \ref{fig:get-review-qbaf}; the updated strengths are depicted next to the arguments (in red and green, respectively).

\subsection{Review QBAFs Combinator} \label{subsec:peerarg-review-qbafs-combinator}

The next step is to combine our extracted review QBAFs.
Review QBAFs share the same aspect arguments and decision argument, but may have different attacks, supports and text arguments. 
Once all the review QBAFs are extracted from the reviews, their text arguments are trimmed off.
Below, 
we define a trimmed QBAF, which simply removes the text arguments from the review QBAF.

\begin{definition}
\label{def:trimmed-review-qbaf}
    Given a review QBAF $\langle X, Att, Supp, \beta \rangle$ under semantics $\sigma$ where $X = T \cup A \cup \{Decision\}$, a trimmed review QBAF is $\langle X_{trim}, Att_{trim}, Supp_{trim}, \beta_{trim} \rangle$ where 
    \begin{align*}
        X_{trim} =&\, A \cup \{Decision\}\\
        Att_{trim} =&\, \{ (b, Decision) | b \in A, (b, Decision) \in Att \}\\
        Supp_{trim} =&\, \{ (b, Decision) | b \in A,(b, Decision)\! \in\! Supp \}\\
        \beta_{trim} =&\, \{ (a, \beta(a)) | a \in A \cup \{Decision\} \}
    \end{align*}
    A semantics of the trimmed review QBAF is a function $\sigma_{trim} = \{ (a, \sigma(a)) | a \in A \cup \{Decision\} \}$.
\end{definition}

In the remainder, 
we will refer to trimmed review QBAFs simply as review QBAFs.
Next, we
combine the frameworks. 

\begin{definition}
\label{def:pre-mpaf}
    Given $n$ review QBAFs $Q_1,\dots, Q_n$ where $Q_i=\langle X, Att_i, Supp_i, \beta_{i}\rangle$ for each $i\leq n$, and semantics $\sigma_{1},\dots,\sigma_n$.
    The \emph{pre-Multi-Party Argumentation Framework (pre-MPAF)} is defined as 
    $\langle X, Und, \beta_{vec}\rangle$
    where $$Und = \{ (a, b) \mid (a, b) \in \bigcup_{i = 1}^{n} (Att_i \cup Supp_i) \}$$
    and $\beta_{vec}: X \rightarrow [0, 1]^{n}$ is a total function such that $\beta_{vec}(a) = [\sigma_{1}(a),\ldots,\sigma_{n}(a)]$ for each argument $a \in X$.
\end{definition}
The outcome of the semantics $\sigma_i$ from the $n$ review QBAFs is the base score of the pre-MPAF. 

\begin{example}
    An example of a pre-MPAF is illustrated below.
    \begin{center}
        \includegraphics[width=1.0\linewidth, keepaspectratio]{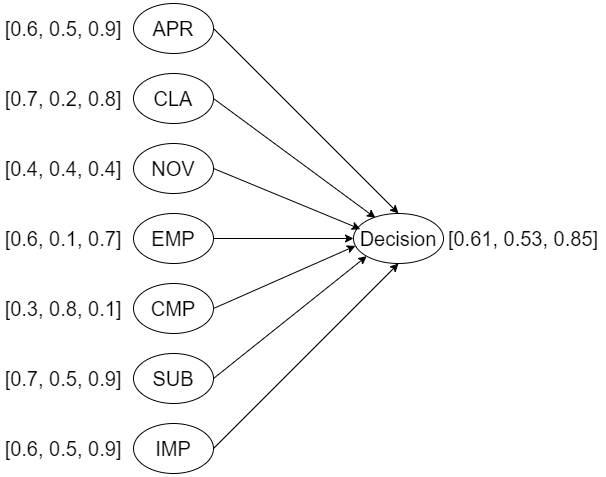}
    \end{center}
    We have $\beta_{vec} = $ \{ (APR, [0.6, 0.5, 0.9]), (CLA, [0.7, 0.2, 0.8]), (NOV, [0.4, 0.4, 0.4]), (EMP, [0.6, 0.1, 0.7]), (CMP, [0.3, 0.8, 0.1]), (SUB, [0.7, 0.5, 0.9]), (IMP, [0.6, 0.5, 0.9]), (Decision, [0.61, 0.53, 0.85]) \}.
\end{example}

\subsection{Pre-MPAF Aggregator} \label{subsec:peerarg-pre-mpaf-aggregator}

\begin{figure}[t]
    \centering
    
    \begin{tikzpicture}[
    ellipsenode/.style={ellipse, draw=green!60, fill=green!5, very thick, minimum size=7mm, font=\small},
    decisionnode/.style={ellipse, draw=blue!60, fill=blue!5, very thick, minimum size=5mm, font=\small},
    squarednode/.style={rectangle, draw=red!60, fill=red!5, very thick, minimum size=5mm, font=\small},
    ]
    
    \node[ellipsenode][align=center] (preMpaf) at (0,0) {pre-MPAF};
    \node[squarednode][align=center] (aggSem1) at (0, -3) {Aggregation Semantics \\ Type 1};
    \node[squarednode][align=center] (aggSem2) [right = of aggSem1] {Aggregation Semantics \\ Type 2};
    \node[squarednode][align=center] (decStrInt) [above = of aggSem2] {Decision Strength \\ Interpretation};
    \node[ellipsenode][align=center] (mpaf) at (0, -4.5) {MPAF};
    \node[squarednode][align=center] (threshold) at (0, -5.5) {Decision Argument's Strength};
    \node[decisionnode][align=center] (accept) at (0, -7) {Accept};
    \node[decisionnode][align=center] (reject) [right = of accept] {Reject};
    
    \draw[->] (preMpaf.south) -- (aggSem1.north) node[midway, left, font=\small] {(1)};
    \draw[->] (preMpaf.south east) -- (decStrInt.north) node[midway, above right, font=\small] {(2)};
    \draw[->] (decStrInt.south) -- (aggSem2.north);
    \draw[->] (aggSem1.south) -- (mpaf.north);
    \draw[->] (aggSem2.south) -- (threshold.north east);
    \draw[->] (mpaf.south) -- (threshold.north);
    \draw[->] (threshold.south) -- (accept.north); 
    \draw[->] (threshold.south) -- (reject.north west); 
    \end{tikzpicture}

    \caption{Pre-MPAF Aggregator diagram, with (1) and (2) as the first and the second implementation types.}
    \label{fig:pre-mpaf-aggregator-diagram}
\end{figure}
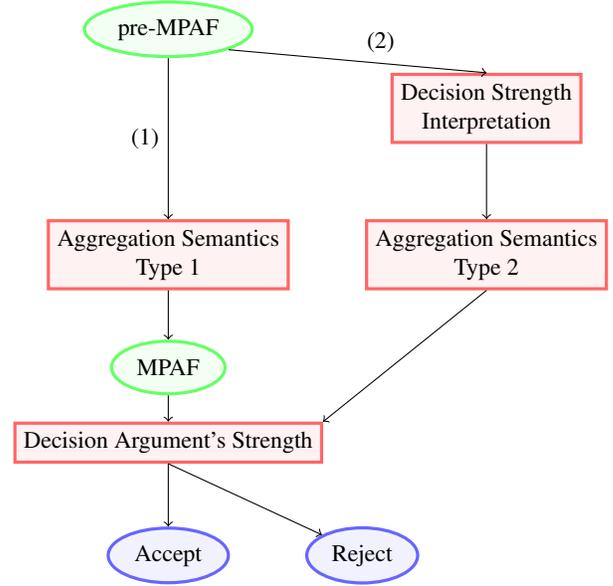

To obtain the final decision, we aggregate the information obtained from the pre-MPAFs.
We implement two types of aggregation methods in PeerArg, as illustrated in Figure~\ref{fig:pre-mpaf-aggregator-diagram}.

The first method (path (1), left, in Figure~\ref{fig:pre-mpaf-aggregator-diagram}) aggregates the strength vectors of the pre-MPAF and uses them to identify support and attack relations. The outcome is a QBAF, called \emph{multi-party argumentation framework (MPAF)}, which is then used to determine the strength of the decision argument, based on DF-QuAD and MLP-based semantics.
In the second method (path (2), right, in Figure~\ref{fig:pre-mpaf-aggregator-diagram}), we focus on the strength of the decision argument. We apply a \emph{decision strength interpretation} to convert a list of strengths of the decision argument into a list of decisions and aggregate them to return the final accept/reject decision.
The final strength of the decision argument $Decision$ is then used to determine the paper acceptance. 
We use a simple threshold such that the paper is predicted to be accepted only if the strength is more than 0.5; otherwise, rejected.

\begin{figure}[t]
    \centering
    \includegraphics[width=1.0\linewidth, keepaspectratio]{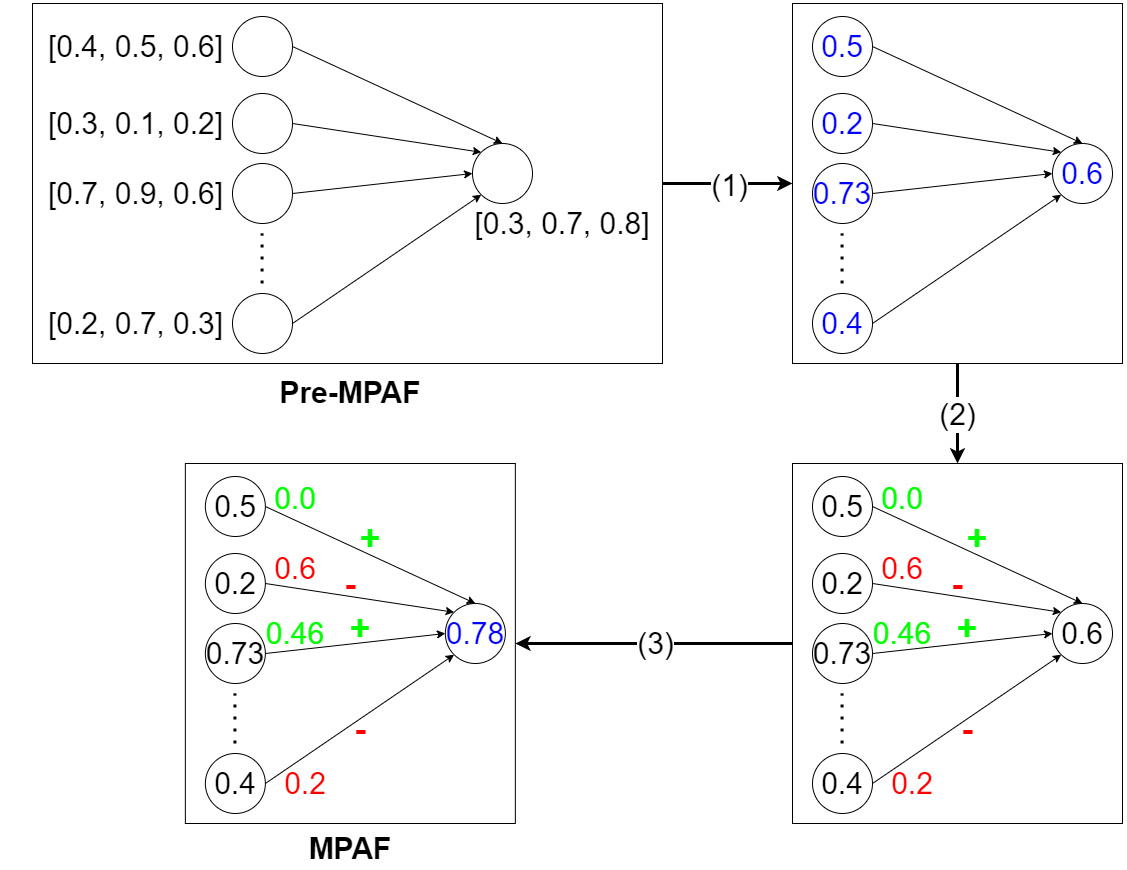}
    \caption{Pre-MPAF to MPAF pipeline. (1) Each strength vector is averaged, giving a strength for the aspect arguments and a base score for the decision argument. (2) Relations and the strength towards the decision arguments are calculated. (3) Semantics (DF-QuAD or MLP-based) is applied to calculate the decision argument's strength.}
    \label{fig:pre-mpaf-to-mpaf}
\end{figure}

\paragraph{Aggregation with Argumentation (Path 1)}
We aggregate our pre-MPAF to obtain an MPAF by applying an aggregation function to identify a strength for each argument and to complete the attack and support relations.
The process is illustrated in Figure \ref{fig:pre-mpaf-to-mpaf}. 
Given a pre-MPAF $\langle X, Und, \beta_{vec}\rangle$, we calculate the average of the sequence of strengths $\beta_{vec}(a)$ for all arguments $a \in X$
to determine the relations between aspect and decision argument(s).
\begin{definition}
\label{def:mpaf}
    Given a pre-MPAF $\langle X, Und, \beta_{vec}\rangle$.
    For each $a\in X$, we calculate the average $$\gamma(a) =  \frac{1}{\vert \beta_{vec}(a) \vert}\sum_{x\in \beta_{vec}(a)}x.  $$
    We define our MPAF $\langle X, Att, Supp, \beta \rangle$ with
    \begin{align*}
        Att =&\, \{ (a,Decision)\in Und \mid \gamma(a) < 0.5 \} \\
        Supp =&\,  \{ (a,Decision)\in Und \mid \gamma(a)\geq 0.5 \} \\
        \beta(a) =&\, 
        \begin{cases}
            \gamma(a) & \text{if } a=Decision\\
            2\cdot \vert \gamma(a)-0.5\vert & \text{ otherwise}
        \end{cases}
    \end{align*}
\end{definition}
We obtain the base score $\beta(a)$ of the arguments by averaging their strength vectors $\beta_{vec}(a)$ and recenter it around 0.5, similar as described in Section \ref{subsec:peerarg-review-qbaf-extractor}. The average $\gamma(a)$ furthermore determines the relation between the decision and the aspect arguments.

In the next step, the DF-QuAD semantics and the MLP-based semantics, respectively, can be applied to calculate the strength of the decision argument of the MPAF.

Similar to review QBAFs, we depict the initial strength of the aspect arguments of MPAFs in the nodes in Figure~\ref{fig:pre-mpaf-to-mpaf}; the updated base scores are depicted next to the arguments. 

\paragraph{Decision Vector Aggregation (Path 2)}
\label{subsubsec:decision-interpret-method}
We convert the strength vector of the decision argument into a decision vector. 
We employ two different decision interpretations: \emph{binary} and \emph{uniform five-level}. 
Binary interpretation simply treats a strength of 0.5 or below to be \emph{'reject'}, and over 0.5 to be \emph{'accept'}. 
For the uniform five-level interpretation, the argument's strength range between 0 and 1 is divided equally into five regions. For a decision argument with strength $s$, the decision $d(s)$ is
\[  
    d(s) = 
    \begin{cases} 
      \text{strong reject} & 0 \leq s < 0.2 \\
      \text{weak reject} & 0.2 \leq s < 0.4 \\
      \text{borderline} & 0.4 \leq s < 0.6 \\
      \text{weak accept} & 0.6 \leq s < 0.8 \\
      \text{strong accept} & 0.8 \leq s \leq 1.0 
   \end{cases}
\]

We obtain a vector where each entry reflects the decision of one individual reviewer.
In the next step, we aggregate the list of decisions.
For this, we use majority-voting and all-accept aggregation, and obtain four aggregation functions: \emph{binary majority-voting}, \emph{uniform five-level majority-voting}, \emph{binary all-accept}, and \emph{uniform five-level all-accept}.

Aggregations based on binary decisions are given below.
\begin{definition}
Let $\vec{d} \in \{reject,accept\}^{k}$ denote a binary decision vector of length $k$, let $Acc = \{d \in \vec{d} \mid d = accept\}$ and $Rej = \{d \in \vec{d} \mid d = reject\}$.
We define two aggregation methods.
\begin{align*}
    \sigma_{\operatorname{binary-majority}}(\vec{d}) = &\,
        \begin{cases}
            1.0 & |Acc| > |Rej| \\
            0.0 & |Acc| \leq |Rej|
        \end{cases}\\
    \sigma_{\operatorname{binary-all-accept}}(\vec{d}) = &\,
        \begin{cases}
            1.0 & |Rej| = 0\\
            0.0 & otherwise
        \end{cases}
\end{align*}
\end{definition}
We define similar aggregation methods for the uniform five-level decision interpretations.
For this, we consider numerical values for the decisions, as usual.
$$
map(d)=
\begin{cases}
      -2 & \text{if } d=\text{strong reject (sr)}\\
      -1 & \text{if } d=\text{weak reject (wr)} \\
      0 & \text{if } d=\text{borderline (bo)}  \\
      1 & \text{if } d=\text{weak accept (wa)} \\
      2& \text{if } d=\text{strong accept (sa)}  
\end{cases}$$
\begin{definition}
Let $\vec{d} \in \{sr,wr,bo,wa,sa\}^{k}$ denote a 5-level decision vector of length $k$.
We define the following aggregation methods. 
\begin{align*}
    \sigma_{\operatorname{5-level-majority}}(\vec{d}) = &\,
        \begin{cases}
            1.0 & \sum_{d\in \vec{d}} map(d) > 0 \\
            0.0 & \sum_{d\in \vec{d}} map(d) \leq 0
        \end{cases}\\
    \sigma_{\operatorname{5-level-all-accept}}(\vec{d}) = &\,
        \begin{cases}
            1.0 & map(d) > 0 \text{ for all }d\in \vec{d} \\
            0.0 & otherwise
        \end{cases}
\end{align*}
\end{definition}
To reach a decision, we apply these functions to the vector associated with the \emph{Decision} argument.
For instance, the binary majority-voting aggregation function converts strengths of the \emph{Decision} argument in the pre-MPAF to 'accept' or 'reject' decisions. The majority decision is then taken, favouring 'reject' if tied. For example, a strength vector [0.1, 0.2, 0.8] would be converted to ['reject', 'reject', 'accept'] and since there are more 'reject' than 'accept', the overall decision is 'reject' and the strength of the decision argument is set to 0.0.

The idea behind the uniform five-level majority-voting aggregation function is to convert each decision strength of the decision argument of the given pre-MPAF into one of the five decisions (strong-reject, weak-reject, borderline, weak-accept, and strong-accept), map them into a weight value from -2 to 2 in order, then sum the total weights. The paper is predicted to be accepted (by setting strength to 1.0) only if the sum is positive. Using the same example as before, the strength vector [0.1, 0.2, 0.8] would be converted to ['strong-reject', 'weak-reject', 'strong-accept'] which is then mapped to [-2, -1, +2]. The sum is -1 which is negative so the strength of the decision argument is set to 0.0.

Instead of counting numbers of 'accept' and 'reject', we predict 'reject' if any decision in a vector is 'reject' for the binary all-accept function.
Similarly, for the uniform five-level all-accept function, the paper is predicted 'accept' only if all the decisions in a vector are strong/weak-accept.

\section{Experiments} \label{sec:experiments}

We evaluate the performance of 
PeerArg in paper acceptance prediction in comparison with the end-2-end LLM approach.
We use three datasets for our classification evaluation: Peer-Review-Analyze (\texttt{PRA}), \texttt{PeerRead}, and Multi-disciplinary Open Peer Review Dataset (\texttt{MOPRD}).
The PRA dataset \citep{PEER-REVIEW-ANALYZE} contains reviews of accepted and rejected papers from ICLR 2018\footnote{The 6th International Conference on Learning Representations; https://iclr.cc/archive/www/2018.html} conference. Each sentence in a review is annotated for which aspects it belongs to, and the sentiments the sentence has towards such aspects.
The PeerRead dataset \citep{PEERREAD} contains reviews from various computer science conferences. In this paper, only the reviews from the ACL 2017 conference were used since their corresponding papers were classified as accepted or rejected. Additionally, each review has scores for each aspect. We therefore consider two cases when these scores are set and not set as base scores of the aspect arguments.
Finally, the MOPRD dataset \citep{MOPRD} contains reviews from several journals in various fields such as computer science, biology, and medicine. In this paper, we used the reviews from the medical field in our evaluation.
\begin{table}[t]
    \centering
    \small 
    \begin{tabular}{ c c c c  } \toprule
    \multirow{2}{4em}{Base \\ Score} & 
    \multirow{2}{4.5em}{QBAF \\ Semantics} & 
    \multirow{2}{3.5em}{Decision \\ Strength } & 
    \multirow{2}{5em}{Aggregation \\ Semantics} \\
    & & & \\ 
    \midrule
    \multirow{2}{4em}{default, \\ sentiment 
    } & 
    \multirow{2}{5em}{DF-QuAD, \\ MLP 
    } & 
    \multirow{2}{3.5em}{binary, \\ 
    5-level} & 
    \multirow{4}{5em}{majority, 
    \\ all-accept, \\ DF-QuAD, \\ MLP} \\ 
    & & & \\ 
    & & & \\
    & & & \\
    \bottomrule
    \end{tabular}
    \caption{PeerArg Hyperparameters}
    \label{tab:hyperparams}
\end{table}

\paragraph{Hyperparameters}
From Section \ref{sec:peerarg-details}, there are 6 possible variables in the PeerArg pipeline that can be adjusted including \emph{aspect classification}, \emph{sentiment analysis}, \emph{base score setting}, \emph{QBAF semantics}, \emph{decision strength interpretation}, and \emph{aggregation semantics}.
We set the aspect classification to be a few-shot learning LLM and the sentiment analysis to be a pretrained sentiment analysis model. The technical details are in Appendices \ref{appendix-sub:aspect-classification} and \ref{appendix-sub:sentiment-analysis}. Accordingly, the remaining four variables are the hyperparameters in our experiments.

All hyperparameter combinations are given in Table \ref{tab:hyperparams}. 
The base score setting determines how the base scores of the text arguments should be set, either to default (0.5) or to the sentiment strengths obtained during the sentiment analysis process. 
For QBAF semantics, we consider either DF-QuAD or MLP-based semantics. 
The decision strength interpretation is either binary or uniform five-level (this only applies for path 2 in Figure~\ref{fig:pre-mpaf-aggregator-diagram}).
Finally, we aggregate using either path 1 in Figure~\ref{fig:pre-mpaf-aggregator-diagram}, i.e., constructing an MPAF and applying DF-QuAD or MLP-based semantics; 
or path 2 in Figure~\ref{fig:pre-mpaf-aggregator-diagram} and aggregate using majority-voting or all-accept.

\paragraph{Experimental Results}
\begin{table}[t]
    \centering
    \setlength{\tabcolsep}{4pt} 
    \small 
    \begin{tabular}{ c c c c c }
    \toprule
    \multirow{2}{4em}{} &
    \multirow{2}{4em}{Base \\ Score} & 
    \multirow{2}{4.5em}{QBAF \\ Semantics} & 
    \multirow{2}{3.5em}{Decision \\ Strength} & 
    \multirow{2}{5em}{Aggregation \\ Semantics} \\
    & & & & \\
    \midrule
    (1) & sentiment & MLP & binary & majority \\
    (2) & sentiment & DF-QuAD & - & DF-QuAD \\ 
    (3) & default & DF-QuAD & 5-level & all-accept \\ 
    (4) & sentiment & DF-QuAD & 5-level & majority \\
    \bottomrule
    \end{tabular}
    \caption{Best PeerArg Hyperparameter Combinations.
    (1) is the best combination for PRA;
    (2) is the best combination for PeerRead with default (0.5) base scores for the aspect arguments;
    (3) is the best combination for PeerRead with reviewer ratings as base scores for the aspect arguments;
    (4) is the best overall combination.
    }
    \label{tab:best_combis}
\end{table}
\begin{figure}[t]
    \centering
    \includegraphics[width=1.0\linewidth, keepaspectratio]{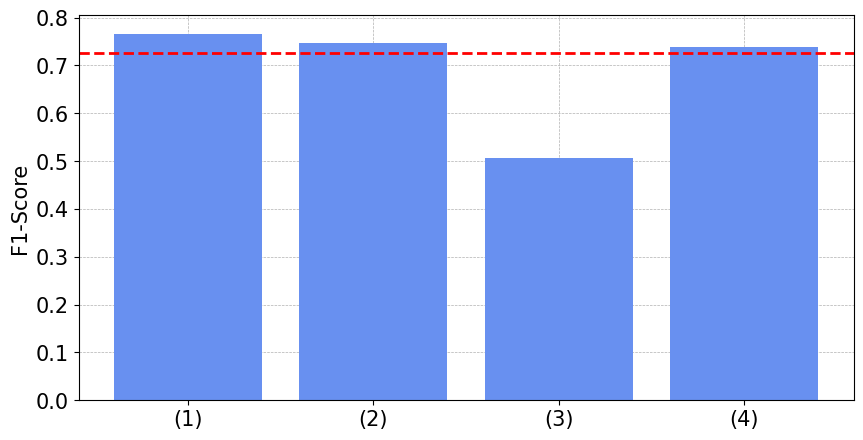}
    \caption{Performances (macro F1) on PRA. 
    The proportion of the correctly predicted paper acceptance is 
    (1) 76.64\%,
    (2) 74.69\%,
    (3) 50.69\%, and
    (4) 73.82\%.
    The red dotted line is the performance of the end-2-end-LLM (72.5\%).}
    \label{fig:pra-chart}
\end{figure}

\begin{figure}[t]
    \centering
    \includegraphics[width=1.0\linewidth, keepaspectratio]{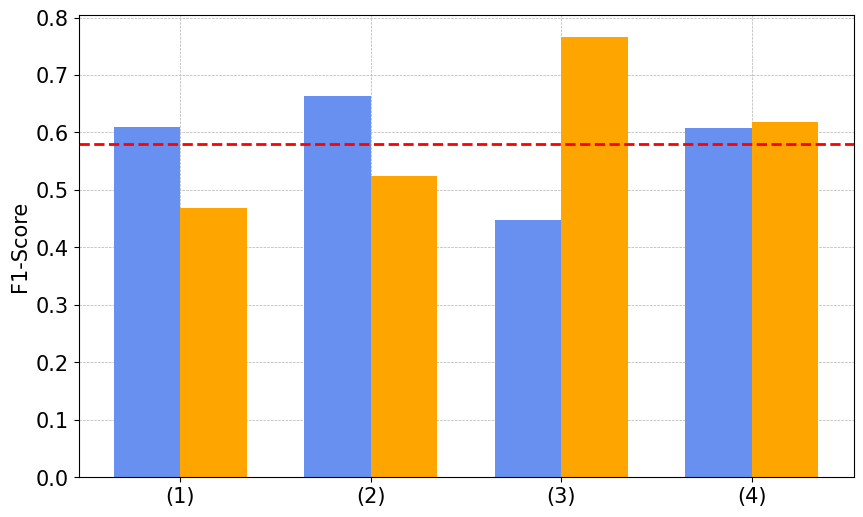}
    \caption{Performances (macro F1) on PeerRead when aspect arguments have default base scores (blue), and when they take on aspect scores from reviews (orange).
    The proportion of the correctly predicted paper acceptance is
    (1) 60.91\% (blue); 46.94\% (orange),
    (2) 66.29\% (blue); 52.49\% (orange),
    (3) 44.83\% (blue); 76.56\% (orange), and
    (4) 60.74\% (blue); 61.87\% (orange),
    The red dotted line is the performance of the end-2-end-LLM (58\%).}
    \label{fig:peerread-chart}
\end{figure}

\begin{figure}[t]
    \centering
    \includegraphics[width=1.0\linewidth, keepaspectratio]{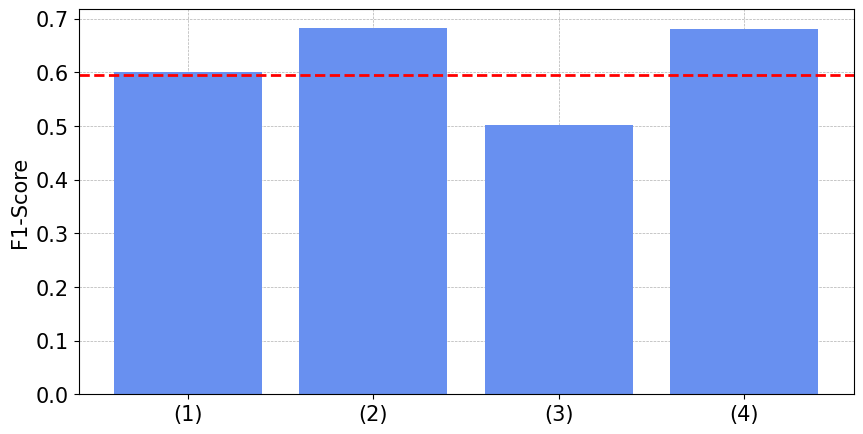}
    \caption{Performances (macro F1) on MOPRD. 
    The proportion of the correctly predicted paper acceptance is 
    (1) 60.12\%,
    (2) 68.32\%,
    (3) 50.27\%, and
    (4) 68.09\%.
    The red dotted line is the performance of the end-2-end LLM (59.5\%).}
    \label{fig:moprd-chart}
\end{figure}

We tested the end-2-end LLM as well as all possible combinations of hyperparameters on PRA and PeerRead datasets, identifying the four best combinations, and testing them along with the end-2-end LLM on the MOPRD.  
Our results show that the end-2-end LLM performs well but is outperformed by PeerRead on all datasets w.r.t.\ the best overall combination of hyperparameters.

Table \ref{tab:best_combis} shows the best combinations of the hyperparameters for PRA and PeerRead datasets.
For PRA (1), the base scores of the aspect arguments were set to 0.5 (default scores). We tested PeerRead with default scores (2) and using the given aspect scores in each review as base scores for the aspect arguments (3).
We find that the best combination of hyperparameters, indicated in (4), is to set sentiment strengths as base scores for text arguments, DF-QuAD semantics, uniform five-level as a decision strength interpretation method, and majority-voting for aggregation.

Crucially, the best overall combination (4) outperformed the end-2-end LLM on all considered datasets.
The performances of the four combinations in comparison with the end-2-end LLM on PRA, PeerRead, and MOPRD are visualised in Figures \ref{fig:pra-chart}, \ref{fig:peerread-chart}, and \ref{fig:moprd-chart} respectively.
For evaluation, we use the macro F1-score which takes both precision (fraction of the true positives over the reported positives) and recall (fraction of the true positives over the number of true positives and false negatives) into account.
One remark is that combination (3) only performs well when the aspect scores from reviews are used, otherwise it is outperformed by the end-2-end LLM in all the other settings.

\section{Conclusion}
We introduced two approaches, PeerArg and an end-2-end LLM,
to enhance the peer reviewing process by predicting paper acceptance from reviews. 
In contrast to the end-2-end LLM that uses few-shot learning techniques to predict paper acceptance in a black-box nature, PeerArg adopts both methods from LLM and computational argumentation to support a decision in the peer reviewing process.
Our experimental results show that PeerArg can outperform the end-2-end LLM, while being more transparent due to the interpretable nature of argumentation.

For future work, we plan to leverage the interpretability of the proposed argumentation model to improve the explainability of the (automated) review aggregation process, similar to how argumentation models are recently used to interpret neural networks, especially the multi-layer perceptrons \citep{NN-to-QBAF}. 
Moreover, we aim to combine
text arguments into pre-MPAFs in the reviews aggregation step. It would also be interesting to explore uncertainty in aggregation and how it would affect the acceptance prediction.

\appendix
\section{Appendix (Technical Details)}
\label{appendix:technical-details}
This section outlines all the relevant models involved in the argument mining process (aspect classification \& sentiment analysis) of the PeerArg pipeline.

\subsection{Aspect Classification}
\label{appendix-sub:aspect-classification}
The LLM for aspect classification is a quantised 4-bit pretrained Mistral-7B-v0.1 model from Mistral AI\footnote{https://huggingface.co/mistralai/Mistral-7B-v0.1}. We used few-shot learning method \citep{FEW-SHOT-LEARNING}. Inspired from \citep{LLM-IN-RBAM}, the template contains a primer and a prompt. The primer has a description of seven aspect criteria for a paper review, and ten different samples of review sentences with their corresponding aspects. The prompt contains the review sentence we want to classify the aspect(s) but with no labels. Note that we split the review into sentences using NLTK sentence tokeniser \citep{NLTK}. We also removed the newlines, backward slashes, and leading dashes from each sentence. 

The partial template for the primer \& prompt input to the LLM is shown in Figure \ref{fig:llm-aspect-template}. The aspect criteria are described first, followed by the samples of sentence-to-aspects classification. Each sample starts with \emph{Sentence:}-placeholder followed by a sentence, then a newline with \emph{Aspects:}-placeholder followed by one or more aspects the sentence is associated with. The prompt section has a similar sentence-aspects template, but leaving space after the \emph{Aspects:}-placeholder to let the LLM predict the aspects.

\begin{figure}[t]
    \centering
    \begin{tikzpicture}
        \draw[rounded corners=10pt, line width=0.5mm, purple] (0, -.5) rectangle (8, 13.5);

        \draw[dashed, gray] (0, 0.55) -- (8, 0.55);

        \node[align=left] at (7.25, 0.8) {\textcolor{gray}{Primer}};
        \node[align=left] at (7.25, 0.3) {\textcolor{gray}{Prompt}};

        \node[align=left, text width=7.25cm, font=\small, below right] at (0.25, 13.25) {
            7 aspect criteria for paper review: \\
            - Appropriateness (APR): Does the paper fit with this conference? \\
            - Clarity (CLA): Is the paper well-written and well structured? Is it clear what was done and why? \\
            - Novelty (NOV): How original is the approach? Does the paper break new ground in topic, methodology, or content? \\
            - Empirical and Theoretical Soundness (EMP): Is the approach sound and well-chosen? Are the empirical claims supported by proper experiments? Are the results of the experiments correctly interpreted? Are the arguments in the paper cogent and well-supported? \\
            - Meaningful Comparison (CMP): Is the work adequately compared against existing literature? Are the references adequate? \\
            - Substance (SUB): Does this paper have enough substance, or would it benefit from more ideas or results? \\
            - Impact (IMP): Is the work significant? Does it inspire new ideas or insights which can be impactful to the community? \\
            N.B. Other comments not belonging to any aspect should be classified as OTHER.
        };

        \node[align=left, text width=7.25cm, font=\small, below right] at (0.25, 4.25) {
            Sentence: -Deep Generative Replay section and description of DGDMN are written poorly and is very incomprehensible. \\
            Aspects: CLA
        };

        \node[align=left, text width=7.25cm, font=\small, below right] at (0.25, 2.65) {
            Sentence: * One observation not discussed by the authors is that the performance of the student network under each low precision regime doesn't improve with deeper teacher networks (see Table 1, 2 \& 3). \\
            Aspects: SUB, EMP

        };

        \node[align=left, text width=7.25cm, font=\small, below right] at (0.25, 0.55) {
            Sentence: \textcolor{red}{input sentence from a review} \\
            Aspects: 
        };
        
    \end{tikzpicture}
    \caption{LLM Aspect Classification Input Template}
    \label{fig:llm-aspect-template}
\end{figure}

\subsection{Sentiment Analysis}
\label{appendix-sub:sentiment-analysis}
Sentiment analysis is done per sentence in a review using a pretrained RoBERTa model trained on Twitter tweets fine-tuned for sentiment analysis\footnote{https://huggingface.co/cardiffnlp/twitter-roberta-base-sentiment-latest}. The model takes an input text and returns an output in the form \emph{$<$label, strength$>$} where \emph{label} is the predicted sentiment (positive/neutral/negative) and \emph{strength} is how likely this sentiment is.

\section*{Acknowledgments}
This research was partially funded by the European Research Council (ERC) under the
European Union’s Horizon 2020 research and innovation programme (grant agreement
No. 101020934, ADIX), by J.P.\ Morgan and by the Royal Academy of Engineering
under the Research Chairs and Senior Research Fellowships scheme, and by the INDICATE project
(Grant No. EP/Y017749/1).

\end{document}